\documentclass{article}
\usepackage[utf8]{inputenc} 
\usepackage{main_nips_w}

\title{ADASECANT: Robust Adaptive Secant Method for Stochastic Gradient}

\author{
Caglar Gulcehre \\
Universit\'e de Montr\'eal \\
\texttt{caglar.gulcehre@umontreal.ca}
\And
Marcin Moczulski \\
University of Oxford \\
\texttt{marcin.moczulski@stcatz.ox.ac.uk}
\AND
Yoshua Bengio \\
Universit\'e de Montr\'eal \\
\texttt{bengioy@iro.umontreal.ca} 
}

\nipsfinalcopy 

\begin{document}

\maketitle

\begin{abstract}
    Stochastic gradient algorithms have been the main focus of large-scale learning problems and 
    they led to important successes in machine learning. The convergence of SGD depends on the
    careful choice of learning rate and the amount of the noise in stochastic estimates of the gradients. 
    In this paper, we propose a new adaptive learning rate algorithm, which utilizes curvature information 
    for automatically tuning the learning rates. The information about the element-wise 
    curvature of the loss function is estimated from the local statistics of the stochastic first order
    gradients. We further propose a new variance reduction technique to speed up the convergence. In our preliminary experiments with deep
    neural networks, we obtained better performance compared to the popular stochastic gradient algorithms.
\end{abstract}

\section{Introduction}

In  this paper we develop a stochastic gradient algorithm that reduces the burden of extensive 
hyper-parameter search for the optimization algorithm. The proposed algorithm exploits
a low variance estimator of curvature of the cost function and uses it to obtain an
automatically tuned adaptive learning rate for each parameter.

In the deep learning and numerical optimization, several papers suggest using a diagonal
approximation of the Hessian (second derivative matrix of the cost function with respect to
parameters), in order to estimate optimal learning rates for stochastic gradient descent over high dimensional
parameter spaces \cite{becker1988improving,schaul2012no,lecun1993automatic}. 
A fundamental advantage of using 
such approximation is
that inverting it is a trivial and cheap operation. 
However generally, for neural networks, the inverse of the diagonal Hessian is usually a bad approximation of the 
diagonal of the inverse of Hessian.
Examples options of obtaining a diagonal approximation of Hessian are the Gauss-Newton matrix \cite{lecun2012efficient} or 
by finite differences \cite{schaul2013adaptive}. Such estimations may however
be sensitive to noise due to the stochastic gradient.  \cite{schaul2012no} suggested a reliable 
way to estimate the local curvature in the stochastic setting by 
keeping track of the variance and average of the gradients.

In this paper, we followed a different approach: instead of using
a diagonal estimate of Hessian, we proposed to estimate curvature along the direction of the
gradient and we apply a new variance reduction technique to compute it reliably. By using root mean square
statistics, the variance of gradients are reduced adaptively with a simple transformation. We keep
track of the estimation of curvature using a technique similar to that proposed by \cite{schaul2012no},
which uses the variability of the expected loss. Standard adaptive learning rate
algorithms only scale the gradients, but regular Newton-like second order methods, can perform
more complicate transformations, e.g. rotating the gradient vector. Newton and quasi-newton
methods can also be invariant to affine transformations in the parameter space. Our proposed
\textbf{Adasecant} algorithm is basically a stochastic rank-$1$ quasi-Newton method. But in comparison with other adaptive
learning algorithms, instead of just scaling the gradient of each parameter, Adasecant can also perform an affine
transformation on them.

\section{Directional Secant Approximation}
\label{sec:dir_sec_approx}

Directional Newton is a method proposed for solving equations with multiple variables
\cite{levin2002directional}.  The advantage of directional Newton method proposed in
\cite{levin2002directional}, compared to Newton's method is that, it does not require 
a matrix inversion and still maintains a quadratic rate of convergence.

In this paper, we developed a second-order directional Newton method for nonlinear optimization.
Step-size $\vt^k$ of update $\Delta^k$ for step $k$ can be written as if it was a diagonal matrix:
\begin{equation}
\label{eqn:dir_newton}
\Delta^k =  \vt^k \odot \nabla_{\TT} \f(\TT^k) = \diag(\vt^k) \nabla_{\TT} \f(\TT^k) = - \diag(\vd^k)(\diag(\H \vd^k))^{-1} \nabla_{\TT} \f(\TT^k)
\end{equation}
where $\TT^k$ is the parameter vector at update $k$, $\f$ is the objective function and $\vd^k$ is a
unit vector of direction that the optimization algorithm should follow.
Denoting by $\vh_i=\nabla_{\TT}\frac{\partial \f(\TT^k)}{\partial \theta_i}$ the $i^{th}$ row of the Hessian matrix $\H$
and by $\nabla_{\TT_i} f(\TT^k)$ the $i^{th}$ element of the gradient vector at update $k$,
a reformulation of Equation \ref{eqn:dir_newton} for each diagonal element of the step-size $\diag(\vt^k)$ is:
 %
 \begin{equation}
\Delta_i^k =
 - t_i^k \nabla_{\TT_i }\f(\TT^k)  =
 - d_i^k \frac{\nabla_{\TT_i }\f(\TT^k)}{\vh^k_i\vd^k}
\text{, so effectively } t_i^k=\frac{d_i^k}{\vh^k_i\vd^k}
\end{equation}
%
%
We can approximate the per-parameter learning rate $t_{i}^k$ following \cite{an2005directional}:
\begin{equation}
\label{eqn:step_size_eq}
t_i^k = \frac{d_i^k}{\vh_i^k \vd^k} \approx \frac{t_i^{k} d_i^k}{\nabla_{\TT_i} \f(\TT^k + \vt^{k} \vd^k) - \nabla_{\TT_i} f(\TT^k)}
\end{equation}
Please note that alternatively one might use the R-op to compute the Hessian-vector product
for the denominator in Equation \ref{eqn:step_size_eq} \cite{schraudolph2002fast}. 

%

To choose a good direction $\vd^k$ in the stochastic setting, we use a block-normalized
gradient vector for each weight matrix $\mW^i_k$ and bias vector $\vb^i_k$ where
$\TT=\left\{\mW^i_k, \vb^i_k\right\}_{i=1\cdots k}$ at each layer $i$ and update $k$,
   i.e. $\vd_{\mW^i_k}^k = \frac{\nabla_{\mW^i_k} \f(\TT)}{||\nabla_{\mW^i_k} \f(\TT)||_2}$ and 
$\vd_k = \left[ \vd_{\mW^0_k}^k \vd_{\vb^0_k}^k \cdots \vd_{\vb^l_k}^k\right]$ for a neural
network with $l$ layers. 
Block normalization of the gradient adds an additional noise, but in practice we did not observe any negative impact of it.
We conjecture that this is due to the angle between the stochastic gradient and the block-normalized gradient still being less than $90$
degrees.
%
%
The update step is defined as $\Delta_{i}^k = t_i^k d_i^k$. The per-parameter learning rate 
$t_i^k$ can be estimated with the finite difference approximation,

\begin{equation}
t_i^k = \frac{\Delta_{i}^k}{\nabla_{\TT_i} \f(\TT^k + \Delta^k) - \nabla_{\TT_i} \f(\TT^k)},
\end{equation}
since, in the vicinity of the quadratic local minima,
\begin{align}
    \nabla_{\TT} \f(\TT^k + \Delta^k) - \nabla_{\TT} \f(\TT^k) \approx \H^{k} \Delta^k
\end{align}
We can therefore recover $\vt^k$ as
\begin{equation}
\vt^k = \diag(\Delta^k)(\diag(\H^{k} \Delta^k))^{-1}.
\end{equation}
The directional secant method basically scales the gradient of each parameter with the curvature along the direction of the gradient vector 
and it is numerically stable.
%
%
\section{Variance Reduction for Robust Stochastic Gradient Descent}
\label{sec:var_reduction}
Variance reduction techniques for stochastic gradient estimators have been well-studied 
in the machine learning literature. Both \cite{wang2013variance} and \cite{johnson2013accelerating} proposed new
ways of dealing with this problem. In this paper, we proposed a new variance reduction technique 
for stochastic gradient descent that relies only on basic statistics related to the gradient. 
Let $\g_i$ refer to the $i^{th}$ element of the gradient vector $\vg$  with respect to the parameters 
$\TT$ and $\E[\cdot]$ be an expectation taken over minibatches and different trajectories of
parameters.

We propose to apply the following transformation to reduce the variance of the stochastic
gradients:
\begin{equation}
\label{eqn:tilde_gamma_i_def}
\tilde{g_i}=\frac{g_i + \gamma_i\E[g_i]}{1+\gamma_i}
\end{equation}
%
%
where $\gamma_i$ is strictly a positive real number. 
Let us note that:
\begin{align}
\E[\tilde{g_i}] = \E[g_i]
 \text{  and }
 \var(\tilde{g_i}) = \frac{1}{(1+\gamma_i)^2}\var(g_i)
\end{align}
%
%
So variance is reduced by a factor of $(1+\gamma_i)^2$ compared to $\var(g_i)$.
%
%
%
%
In practice we do not have access to $\E[g_i]$, therefore 
a biased estimator $\overline{g_i}$ based on past values of $g_i$ will be used instead.
We can rewrite the $\tilde{g_i}$ as: 
\begin{equation}
\tilde{g_i} = \frac{1}{1+\gamma_i}g_i + (1 - \frac{1}{1+\gamma_i}) \E[g_i] 
\end{equation}
After substitution $\beta_i = \frac{1}{1+\gamma_i}$, we will have:
\begin{equation}
 \tilde{g_i} = \beta_i g_i + (1 - \beta_i) \E[g_i]
\end{equation}
By adapting $\gamma_i$ or $\beta_i$, it is possible to control the influence of 
high variance, unbiased $g_i$ and low variance, biased $\overline{g_i}$ on $\tilde{g_i}$.
Denoting by $\vg^{\prime}$ the stochastic gradient obtained on the next minibatch, 
the $\gamma_i$ that well balances those two influences is the one that keeps the $\tilde{g_i}$ 
as close as possible to the true gradient $\E[g_i^{\prime}]$ with $g_i^{\prime}$ being the only sample of 
$\E[g_i^{\prime}]$ available:
%
\begin{equation}
\label{eqn:exp_gamma_cri}
\argmin_{\beta_i} \E[||\tilde{g_i} - g_i^{\prime}||^2_2]
\end{equation}
It can be shown that this a convex problem in $\beta_i$ with a closed-form solution (details in
appendix) and we can obtain the $\gamma_i$ from it:
%
%
%
%
%
%
\begin{equation}
\label{eqn:gamma_i_formula}
\gamma_i = \frac{\E[(g_i - g_i^{\prime})(g_i - \E[g_i])]}{\E[(g_i-\E[g_i])(g^{_i\prime}-\E[g_i]))]}
\end{equation}
As a result, in order to estimate $\gamma$ for each dimension, we keep track of a estimation of $\frac{\E[(g_i - g_i^{\prime})(g_i - \E[g_i])]}{\E[(g_i - \E[g_i])(g_i^{\prime}-\E[g_i]))]}$
during training. The necessary and sufficient condition here, for the variance reduction is to
keep $\gamma$ positive, to achieve a positive estimate of $\gamma$ we used the root mean square
statistics for the expectations.
\section{Adaptive Step-size in Stochastic Case}

In the stochastic gradient case, the step-size of the directional secant can be computed by using an expectation over the minibatches:
\begin{equation}
\label{eqn:adapt_stepsize}
\E_k[t_{i}] = \E_k[\frac{\Delta_i^k}{\nabla_{\TT_i} \f(\TT^k +  \Delta^k) - \nabla_{\TT_i} \f(\TT^k)}]
\end{equation}
%
The $E_k[\cdot]$ that is used to compute the secant update, is taken over the minibatches at the past
values of the parameters.

Computing the expectation in Eq \ref{eqn:adapt_stepsize} was numerically unstable in stochastic
setting. We decided to use a more stable second order Taylor approximation of 
Equation \ref{eqn:adapt_stepsize} around $(\sqrt{\E_k[(\alpha_i^k)^2]}, \sqrt{\E_k[(\Delta_i^k)^2]})$, with $\alpha_i^k = \nabla_{\TT_i} \f(\TT^k + \Delta^k) - \nabla_{\TT_i} \f(\TT^k)$.
Assuming  $\sqrt{\E_k[(\alpha_i^k)^2]} \approx \E_k[\alpha_i^k]$ and $\sqrt{\E_k[(\Delta_i^k)^2]} \approx \E_k[\Delta_i^k]$ we obtain always non-negative approximation of $\E_k[t_{i}]$:
\begin{align}
\label{eqn:step_size_stoc_der1}
& \E_k[t_{i}] \approx \frac{\sqrt{\E_k[(\Delta_i^k)^2]}}{\sqrt{\E_k[(\alpha_i^k)^2]}} -
\frac{\cov(\alpha_i^k,\Delta_i^k)}{\E_k[(\alpha_i^k)^2]} 
\end{align}
In our experiments, we used a simpler approximation, 
which in practice worked as well as formulations in Eq \ref{eqn:step_size_stoc_der1}:
\begin{equation}
\label{eqn:step_size_stoc2}
\E_k[t_{i}] \approx \frac{\sqrt{\E_k[(\Delta_i^k)^2]}}{\sqrt{\E_k[(\alpha_i^k)^2]}} -
\frac{\E_k[\alpha_i^k\Delta_i^k]}{\E_k[(\alpha_i^k)^2]}
\end{equation}
%
%
%
%
%
\section{Algorithmic Details}
\subsection{Approximate Variability}
To compute the moving averages as also adopted by \cite{schaul2012no}, 
we used an algorithm to dynamically decide the time constant based on the step size being taken. 
As a result algorithm that we used will give bigger weights to the updates that have large step-size 
and smaller weights to the updates that have smaller step-size.

By assuming that $\bar{\Delta}_i[k]\approx\E[\Delta_i]_k$, 
the moving average update rule for $\bar{\Delta}_i[k]$ can be written as,
\begin{align}
    \label{eqn:approx_var}
    & \bar{\Delta}_i^2[k] = (1~-~\tau_i^{-1}[k])\bar{\Delta}_i^2[k-1] + \tau_i^{-1}[k](t_i^k \tilde{\vg}_i^k), \text{ and } 
    \bar{\Delta}_i[k] = \sqrt{\bar{\Delta}_i^2[k]}
\end{align}
This rule for each update assigns a different weight to each element of the gradient vector .
At each iteration a scalar multiplication with $\tau_i^{-1}$ is performed and $\tau_i$ is adapted using the following equation:
\begin{equation}
    \label{eqn:approx_var_update}
    \tau_i[k] = (1~-~\frac{\E[\Delta_i]_{k-1}^2}{\E[(\Delta_i)^2]_{k-1}})\tau_i[k-1] + 1
\end{equation}
%
\subsection{Outlier Gradient Detection}
%
Our algorithm is very similar to \cite{schaul2013adaptive}, but instead of
incrementing $\tau_i[t+1]$ when an outlier is detected, the time-constant is reset 
to $2.2$. Note that when $\tau_i[t+1] \approx 2$, this assigns approximately the same amount of
weight to the current and the average of previous observations.  This mechanism made learning more stable, 
because without it outlier gradients saturate $\tau_i$ to a large value.
\subsection{Variance Reduction}
The correction parameters $\gamma_i$ (Eq \ref{eqn:gamma_i_formula}) allows for a fine-grained variance reduction for each parameter independently. 
%
The noise in the stochastic gradient methods can have advantages both in terms of generalization and optimization. 
It introduces an exploration and exploitation trade-off, which can be controlled 
by upper bounding the values of $\gamma_i$ with a value $\rho_i$, so that 
thresholded $\gamma_i^{\prime} = \min(\rho_i, \gamma_i)$.
%

We block-wise normalized the gradients of each weight matrix and bias vectors in $\vg$ to compute the
$\tilde{\vg}$ as described in Section~\ref{sec:dir_sec_approx}. 
That makes Adasecant scale-invariant, thus more robust to the scale of the inputs and the number of the layers of the network.
We observed empirically that it was easier to train very deep neural networks with block normalized gradient descent.
\section{Improving Convergence}
Classical convergence results for SGD are based on the conditions:
\begin{equation}
\sum_i (\eta^{(i)})^2 < \infty \text{ and }   \sum_i \eta^{(i)} = \infty
\end{equation}
such that the learning rate $\eta^{(i)}$ should decrease \cite{robbins1951stochastic}. Due to the noise in the estimation of
adaptive step-sizes for Adasecant, the convergence would not be guaranteed. To ensure it, 
we developed a new variant of Adagrad \cite{duchi2011adaptive} with thresholding, 
such that each scaling factor 
is lower bounded by $1$. Assuming  $a_i^k$ is the accumulated norm of all past gradients for $i^{th}$ parameter at update $k$, it is thresholded from below ensuring that the algorithm will converge:
\begin{equation}
a_i^k = \sqrt{\sum_{j=0}^k (g_i^j)^2} \text{ and } \rho_i^k = \text{maximum}(1, a_i^k) \text{ giving }  \Delta_i^k = \frac{1}{\rho_i}\eta_i^k\tilde{\vg}_i^k
\end{equation}
In the initial stages of training, accumulated norm of the per-parameter gradients can be less than $1$. If the
accumulated per-parameter norm of a gradient is less than $1$, Adagrad will augment 
the learning-rate determined by Adasecant for that update, i.e.
$\frac{\eta_i^k}{\rho_i^k} > \eta_i^k$ where $\eta_i^k=\E_k[t_i^k]$ is the per-parameter learning rate determined
by Adasecant. This behavior tends to create unstabilities during the training
with Adasecant. Our modification of the Adagrad algorithm is to ensure that, it will
 reduce the learning rate determined by the Adasecant algorithm at each update, i.e. $\frac{\eta_i^k}{\rho_i^k} \le
 \eta_i^k$ and the learning rate will be bounded. At the beginning of the training, parameters of a
 neural network can get $0$-valued gradients, e.g. in the existence of dropout and ReLU units. 
 However this phenomena can cause the per-parameter learning rate scaled by Adagrad to be
 unbounded. 

In Algorithm \ref{alg:adasecant}, we provide a simple pseudo-code of the Adasecant algorithm.

\section{Experiments}
\begin{figure}[htbp]
\centering
\subfigure[$2$ layer Maxout Network]{
\label{fig:depthexpa}
\includegraphics[width=7cm]{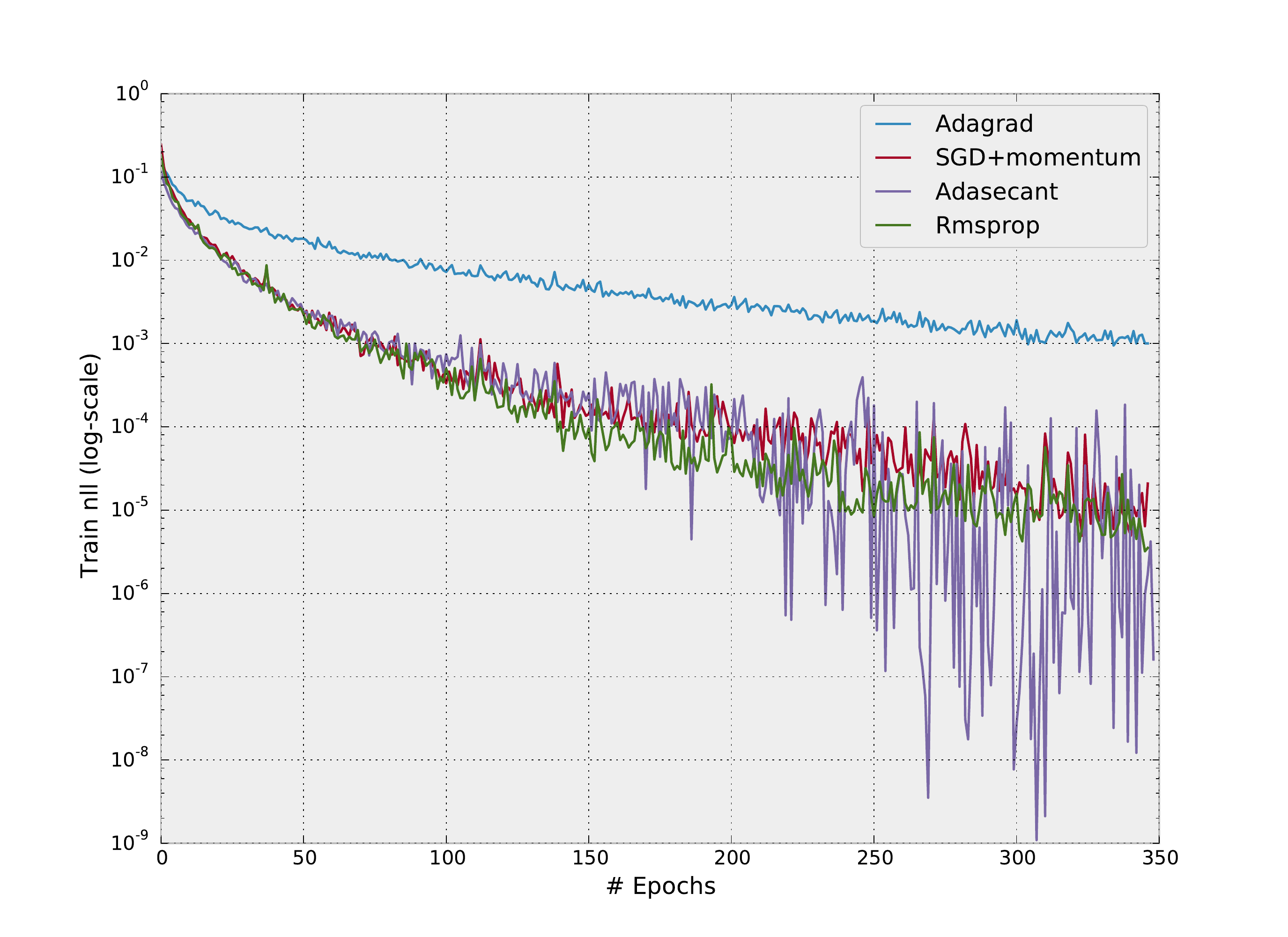}
}
\subfigure[$16$ layer Maxout Network]{
\label{fig:depthexpb}
\includegraphics[width=7cm]{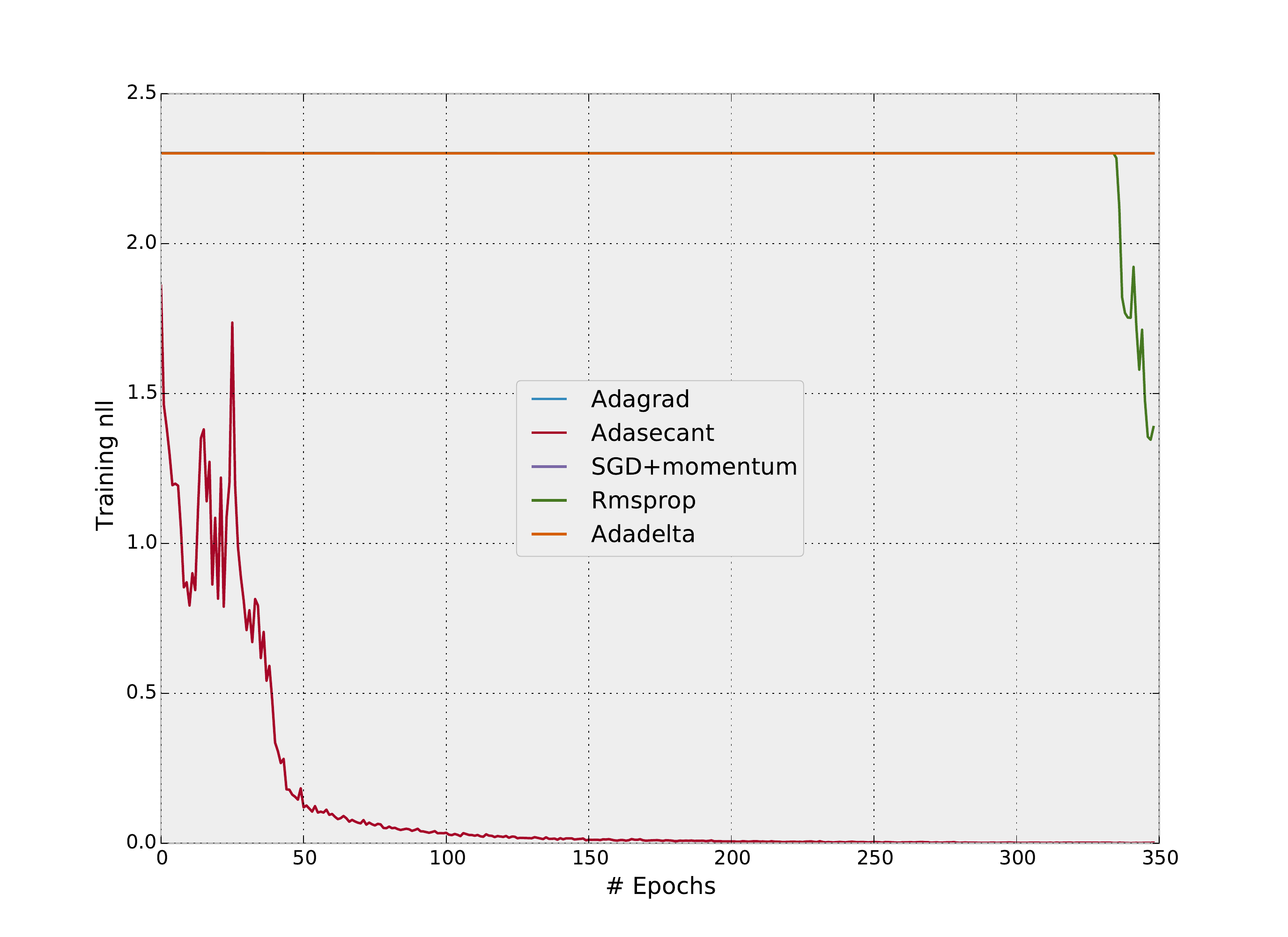}
}
\caption{Comparison of different stochastic gradient algorithms on MNIST with Maxout Networks.
    Both a) and b) are trained with dropout and maximum column norm constraint regularization on
    the weights. Networks are initialized with weights sampled from a Gaussian distribution with
    $0$ mean and standard deviation of $0.05$. In both experiments, the proposed algorithm,
    Adasecant, seems to be converging faster and arrives to a better minima in training set. We trained both
    networks for $350$ epochs over the training set.}
\label{fig:depthexp}
\end{figure}

We have performed our experiments on MNIST with Maxout Networks \cite{goodfellow2013maxout}
comparing Adasecant with popular stochastic gradient learning algorithms: Adagrad, Rmsprop \cite{graves2013generating}, Adadelta \cite{zeiler2012adadelta} and SGD+momentum (with linearly decaying learning rate). 
Results are summarized in Figure~\ref{fig:depthexp} at the appendinx and we showed that Adasecant converges
as fast or faster than other techniques, including the use of hand-tuned global learning rate and momentum for SGD,
RMSprop, and Adagrad.

\section{Conclusion}
We described a new stochastic gradient algorithm with adaptive learning rates
that is fairly insensitive to the tuning of the hyper-parameters and doesn't require tuning of learning
rates. Furthermore, the variance reduction technique we proposed improves the
convergence when the stochastic gradients have high variance. 
According to preliminary experiments presented, we were able to obtain a better
training performance compared to other popular, well-tuned stochastic gradient algorithms. As a future
work, we should do a more comprehensive analysis, which will help us to better understand
the algorithm both analytically and empirically.

\vspace*{-1mm}
\subsubsection*{Acknowledgments}
\vspace*{-1mm}

We thank the developers of Theano \cite{Bastien-2012} and Pylearn2 \cite{pylearn2_arxiv_2013} and the computational resources
provided by Compute Canada and Calcul Qu\'ebec. This work has been partially supported by NSERC,
CIFAR, and Canada Research Chairs, Project TIN2013-41751, grant 2014-SGR-221. We would 
like to thank Tom Schaul for the valuable discussions. We also thank Kyunghyun Cho and Orhan Firat
for proof-reading and giving feedbacks on the paper.

{\small
\bibliography{myref}
\bibliographystyle{plain}}

\appendix
\section{Appendix}
\subsection{Derivation of Eq \ref{eqn:exp_gamma_cri}}

\begin{align}
\label{eqn:exp_gamma_opt2}
\frac{\partial \E[(\beta_i g_i + (1 - \beta_i) \E[g_i] - g_i^{\prime})^2]}{\partial \beta_i}&=0 \\
\E[(\beta_i g_i + (1 - \beta_i) \E[g_i] - g_i^{\prime})\frac{\partial (\beta_i g_i + (1 - \beta_i) \E[g_i] - g_i^{\prime})}{\partial \beta_i}]&=0 \\
\E[(\beta_i g_i + (1 - \beta_i) \E[g_i] - g_i^{\prime})(g_i -  \E[g_i])]&=0 \\
\E[(\beta_i g_i (g_i -  \E[g_i]) + (1 - \beta_i) \E[g_i] (g_i - \E[g_i]) - \E[g_i](g_i - \E[g_i])]&=0 \\
\beta_i = \frac{\E[(g_i - \E[g_i])(g_i^{\prime} - \E[g_i])]}{\E[(g_i - \E[g_i])(g_i - \E[g_i])]} 
= \frac{\E[(g_i - \E[g_i])(g_i^{\prime} - \E[g_i])]}{\var(g_i)} 
\end{align}

\subsection{Algorithm pseudo-code}
Algorithm \ref{alg:adasecant} contains the pseudo-code of the Adasecant algorithm.

\begin{algorithm}[htb]
\SetKwInOut{Input}{input}
\SetKwInOut{Output}{output}
\label{alg:adasecant}
\caption{Adasecant: minibatch-Adasecant for adaptive learning rates with variance reduction}

 \Repeat{stopping criterion is met}{
  draw $\popsize$ samples,
  compute the gradients $\vg^{(j)}$ where $\vg^{(j)} \in \R^{n}$ for each minibatch $j$,
  $\vg^{(j)}$ is computed as,~$\frac{1}{\popsize}\sum_{k=1}^{\popsize}\nabla_{\TT}^{(k)}\f(\TT)$ \\ 
  block-wise normalize gradients of each weight matrix and bias vector \\
  \For{\text{parameter} $i \in \{1, \ldots, n\}$}{
    compute the correction term by using, $\gamma^k_i =\frac{\E[(g_i - g_i^{\prime})(g_i - \E[g_i])]_k}{\E[(g_i-\E[g_i])(g_i^{\prime}-\E[g_i]))]_k}$ \\
  	compute corrected gradients $\tilde{g_i}=\frac{g_i+\mathbf{\gamma_i}\E[g_i]}{1+\mathbf{\gamma_i}}$ \\
  	\vspace{0.5em}
    \If{$|g_i^{(j)}- \E[g_i]| > 2 \sqrt{\E[(g_i)^2]-(\E[g_i])^2} \;\;\; \operatorname{or} \;\;\;
        \left|\alpha_i^{(j)} - \E[\alpha_i]\right| > 2 \sqrt{\E[(\alpha_i)^2] - (\E[\alpha_i])^2}$}
  	{\vspace{0.5em}
  	reset the memory size for outliers $\tau_i \leftarrow 2.2$}{}
  	\vspace{0.5em}
    update moving averages according to Equation \ref{eqn:approx_var} \\
\vspace{0.5em}
  estimate learning rate $\;\;\displaystyle\eta_i^{(j)} \leftarrow
      \frac{\sqrt{\E_k[(\Delta_i^{(k)})^2]}}{\sqrt{\E_k[(\alpha_i^k)^2]}} -
      \frac{\E_k[\alpha_i^k\Delta_i^k]}{\E_k[(\alpha_i^k)^2]}$\\
  \vspace{0.5em}
  update memory size as in Equation \ref{eqn:approx_var_update}\\
 \vspace{0.5em}
 update parameter $\;\; \theta_i^j \leftarrow \theta_i^{j-1} - \eta_i^{(j)}\cdot \tilde{g}_i^{(j)}$\\
 }
 }

\end{algorithm}
\subsection{Further Experimental Details}
In our experiments with Adasecant algorithm, adaptive momentum term $\gamma_i^k$ was clipped at
$1.8$. In $2$-layer Maxout network experiments for SGD-momentum experiments, we used the best hyper-parameters reported 
by \cite{goodfellow2013maxout}, for Rmsprop and Adagrad, we crossvalidated learning rate for $15$ different learning rates 
sampled uniformly from the log-space. We crossvalidated $30$ different pairs of momentum and learning rate for SGD+momentum,
for Rmsprop and Adagrad, we crossvalidated $15$ different learning rates sampled them from log-space uniformly
 for deep maxout experiments.
In Figure~\ref{fig:adasecant_clock}, we analyzed the effect of using
different minibatch sizes for Adasecant and compared its convergence with Adadelta in wall-clock
time. For minibatch size $100$ Adasecant was able to reach the almost same training negative log-likelihood 
as Adadelta
after the same amount of time, but its convergence took much longer. With minibatches of size $500$ Adasecant 
was able to converge faster in wallclock time to a better local minima.

\begin{figure}[htp]
\centering
\label{fig:adasecant_clock}
\includegraphics[width=10cm]{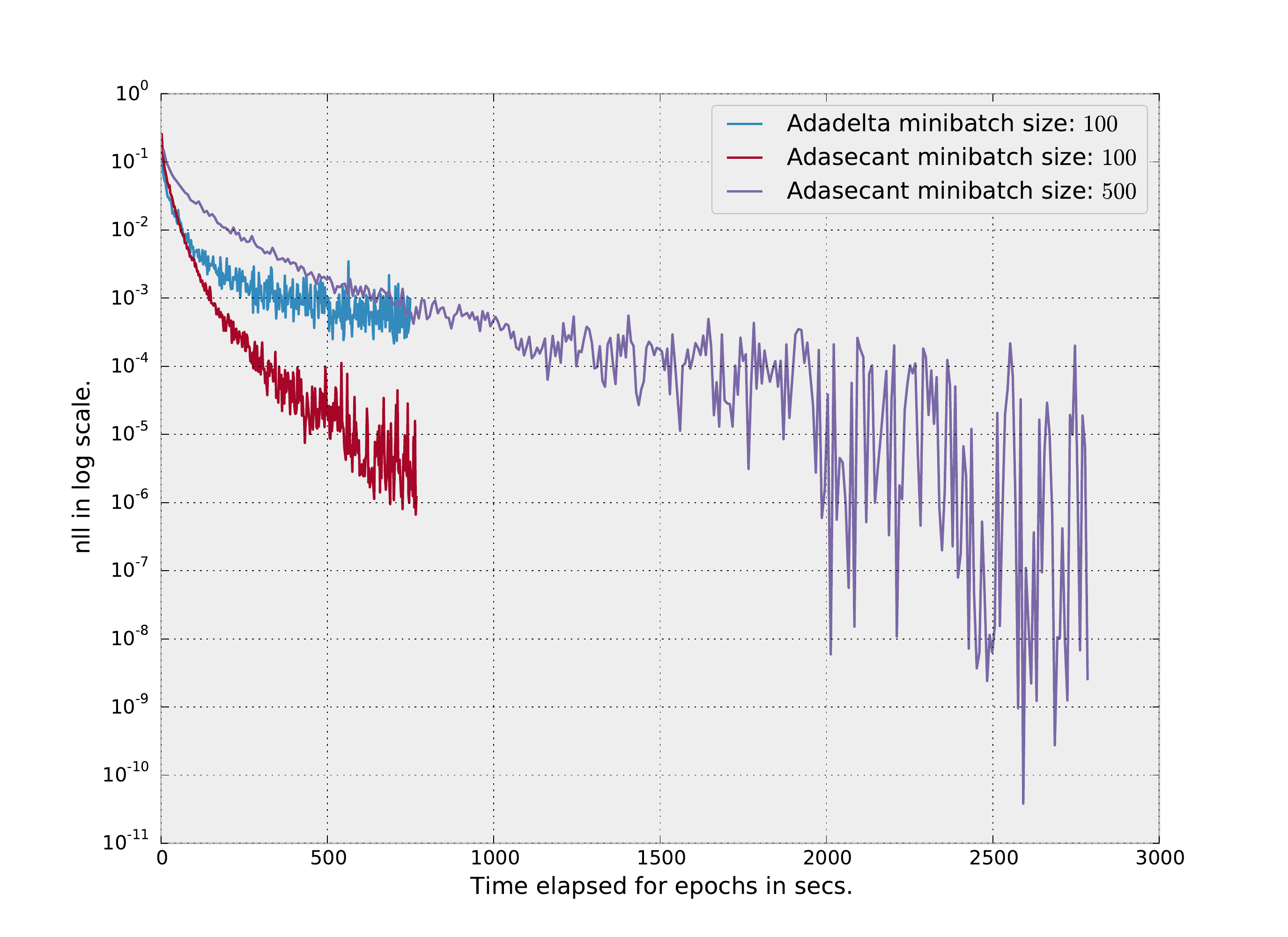}
\caption{In this plot, we compared adasecant trained by using minibatch size of $100$ and $500$
    with adadelta using minibatches of size $100$. We performed these experiments on MNIST with
    2-layer maxout MLP using dropout.}
\label{fig:adasecant_clock}
\end{figure}

\end{document}